\pdfoutput=1
\documentclass[11pt]{article}
\usepackage{ACL2023}

\usepackage{times}
\usepackage{latexsym}

\usepackage[T1]{fontenc}
\usepackage[utf8]{inputenc}
\usepackage{microtype}
\usepackage{inconsolata}
\usepackage{enumitem,kantlipsum}
\usepackage{graphicx}
\usepackage{caption}
\usepackage{amsmath}
\usepackage{amssymb}
\usepackage{pgfplots}
\pgfplotsset{compat=newest}
\usepackage{comment}
\usepackage{booktabs}
\usepackage{hyperref}
\usepackage[overload]{empheq}
\usepackage{makecell}
\usepackage{xcolor,colortbl}
\usepackage[ruled,vlined]{algorithm2e}
\usepackage{array,multirow}
\usepackage{comment}
\usepackage{pifont}
\usepackage{url}

\newcommand{\splitcell}[2][c]{\begin{tabular}[#1]{@{}c@{}}#2\end{tabular}}
\newcommand{\cmark}{\ding{51}}%
\newcommand{\xmark}{\ding{55}}%

\definecolor{Gray}{gray}{0.95}
\newcolumntype{g}{>{\columncolor{Gray}}c}

\title{Reliable and Interpretable Drift Detection in Streams of Short Texts}

\author{
	Ella Rabinovich\hspace{0.85cm}
	Matan Vetzler\hspace{0.85cm}
	Samuel Ackerman\hspace{0.85cm}
	Ateret Anaby-Tavor \\
	\vspace{0.05cm} \\
	IBM Research \\
	\texttt{\{ella.rabinovich1, matan.vetzler, samuel.ackerman\}@ibm.com} \\
	\texttt{atereta@il.ibm.com}
}

\begin{document}
\maketitle
\begin{abstract}
Data drift is the change in model input data that is one of the key factors leading to machine learning models performance degradation over time. Monitoring drift helps detecting these issues and preventing their harmful consequences. Meaningful drift interpretation is a fundamental step towards effective re-training of the model. 
In this study we propose an end-to-end framework for reliable model-agnostic change-point detection and interpretation in large task-oriented dialog systems, proven effective in multiple customer deployments. We evaluate our approach and demonstrate its benefits with a novel variant of intent classification training dataset, simulating customer requests to a dialog system. We make the data publicly available. 

\end{abstract}

\section{Introduction}

Contemporary data centers rely heavily on machine learning services in their deployed systems. These systems are vulnerable to the data drift problem: the phenomenon where the statistical properties of the underlying independent variable change over time. As a concrete example, consider the case where the distribution of data arriving to a supervised classifier gradually diverges from that the model was trained on. Such a phenomenon introduces one of the key challenges in maintaining large models, where drift typically results in performance degradation. Manual inspection of the data is labor-intensive and error-prone, and actual drift might remain unnoticed. Automatic monitoring and detection of divergences in incoming data streams facilitates early risk mitigation introduced by drift.

Goal-oriented dialog systems\footnote{Also referred to as "task-oriented" dialog systems, or "virtual assistants" (VA).} have gained much attention in both the academic and industrial communities over the past decade. The core component of a task-oriented dialog system is the NLU module: the user utterance is either transformed into a modeled intent\footnote{An "intent" is the general topic label value under which user utterances fall, and is identified by a pre-trained intent classifier.  For instance, utterances like "reset login" and "I lost my password" fall under the intent label "account password".} with an appropriate flow of subsequent actions, or labeled as unrecognized and stored in the \textit{unhandled pool} of out-of-scope requests. In practice, the NLU module makes use of a supervised text classifier, where data drift is triggered by "production" data (customer queries) that changes away from the distribution the classifier was trained on. Here we address the scenario of data drift detection in the context of large deployments of task-oriented dialog systems, where emergence of novel topics or deviations in the way customer introduce queries is not uncommon.

Existing approaches to data drift detection are roughly categorized across two functional dimensions: (1) model-dependent vs. model-agnostic and (2) anomaly detection vs. change point detection. In the context of first dimension, `model' refers specifically to a predictive model (e.g., classifier) receiving the text stream as inputs.  A \textit{model-dependent} method directly considers the underlying model performance to detect drift, e.g., by tracking the posterior probability estimates of a classifier.  A \textit{model-agnostic} method uses only the incoming data itself for change detection, e.g., by measuring changes in text representations, whether or not such changes ultimately would affect the performance of a classifier model trained on anchor data. 
Another strength of the model-agnostic approach lies in its direct access to data: once detected, the drift can be explained and interpreted, thereby potentially leading to actionable recommendations. 

\begin{figure*}[h!]
\centering
\resizebox{0.95\textwidth}{!}{
\includegraphics{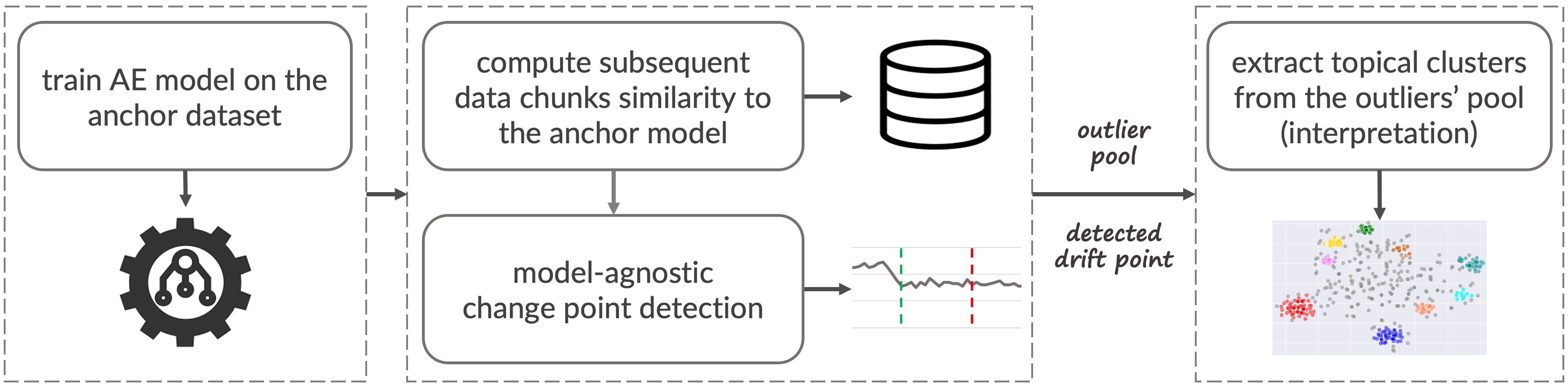}
}
\caption{The end-to-end pipeline for drift detection and interpretation: (1) An autoencoder (AE) is trained on the anchor dataset, reliably representing data distribution at the model training point; (2) each newly arriving batch of requests is examined by the AE, yielding its overall similarity to the anchor dataset, while identifying a set of outliers; (3) change-point detection module is applied on the growing list of similarity observations; and finally (4) topical clustering is applied on a subset of the outlier pool, in case drift is detected.}
\label{fig:process-overview}
\end{figure*}

The second dimension distinguishes between \textit{anomaly} detection that identifies outliers at the single chunk level, and \textit{change-point} detection where a window of recent samples is examined to detect (with statistical guarantees) a point at time where the underlying data distribution undergoes a change; the latter models are robust to noise and transient changes. We propose a pipeline for model-agnostic change-point detection in task-oriented dialog systems. 
Figure~\ref{fig:process-overview} illustrates the pipeline, which is further described in detail in Section~\ref{sec:model}.

User requests towards a dialog system---natural language utterances at the first point of interaction---often contain personal and sensitive information, and are subject to agreements that prevent providers from sharing this data publicly. Extending a large, diverse and publicly-available intent classification dataset \citep{larson-etal-2019-evaluation}, we build a corpus that closely resembles a dialog system request stream, further using it for evaluation of the proposed drift detection approach.
\footnote{The dataset is available at \url{https://huggingface.co/datasets/ibm/clinic150-sur}.}

The contribution of this work is, therefore, two-fold. First, we propose and evaluate an end-to-end pipeline for model-agnostic change-point detection in task-oriented dialog systems, that has been proven effective in multiple large-scale customer deployments. Second, we create and release a extension of an intent classification training dataset that closely imitates the nature of streaming requests towards a virtual assistant.

\section{Related Work}

The main approaches to drift detection in textual streams are model-dependent: they rely on the performance of the underlying classification model, where decreasing classifier confidence (or increasing error rate) are indicative of divergence in statistical properties of the data the classifier was trained on, and the newly arriving texts \citep{ryu2012efficient, sethi2017reliable, ackerman2021automatically}. 
Model-agnostic (e.g., classifier-independent) approaches are commonly applied for novelty detection in textual streams that do not necessarily undergo  classification, like news or a tweet feed. As an example, \citet{spinosa2007olindda} and \citet{hayat2010dct} use concept-based clusters to represent data distributions of temporal textual chunks, and then detect the hidden topic drifts in terms of the difference between concept-based clusters in two adjoining data chunks. A similar approach for novelty detection in textual data was also applied by \citet{faria2013novelty} and \citet{li2017learning}.

Both the model-aware and model-agnostic approaches are commonly used to detect a point of change, regardless of the subsequent trend of the introduced novelty. As such, these methodologies are best associated with "anomaly" or "outlier" detection, while in the case of a task-oriented dialog system we are interested in detecting a systematic, consistent drift trend -- a potential trigger for intent classifier retraining. The only study addressing change point detection in textual data that we are aware of is \citet{wang2018real}, who apply LDA \citep{blei2003latent} for detecting change points in document streams from twitter and news feed. While LDA can be used effectively for modeling long documents, it is practically inapplicable for short (often 2--3 word) requests.
Table \ref{tbl:related-work} summarizes the landscape of the prior art in the domain of drift detection in textual streams. Our work bridges the gap in the domain of \textit{model-agnostic}, statistically-robust \textit{change-point} detection for streams of short texts, while interpreting the detected drift. 

\begin{table}[h!]
\centering
\resizebox{\columnwidth}{!}{
\begin{tabular}{l|cgcg}
study & OD & CPD & M-AW & M-AG \\ \hline
\citet{ryu2012efficient} & \cmark & & \cmark & \\
\citet{sethi2017reliable} & \cmark & & \cmark &  \\
\citet{ackerman2021automatically} & & \cmark & \cmark &  \\
\citet{hayat2010dct} & \cmark & & & \cmark \\
\citet{faria2013novelty} & \cmark & & & \cmark \\
\citet{li2017learning} & \cmark & & & \cmark \\
\citet{spinosa2007olindda} & \cmark & & & \cmark \\ 
\citet{wang2018real} & & \cmark & & \cmark \\ \hline
our study (short texts) & & \cmark & & \cmark \\

\end{tabular}
}
\caption{Representative landscape of prior art in the domain of drift detection. "OD", "CPD", "M-AW" and "M-AG" denote outlier detection, change-point detection, model-aware and model-agnostic, respectively. The LDA-based approach by \citet{wang2018real} is only applicable to document-length texts.}
\label{tbl:related-work}
\end{table}

\section{Dataset}

We study the phenomenon of data drift in the context of (natural-language) user requests to a task-oriented dialog system. Novel topics, or deviations from existing topics can emerge as the result of new services introduced by a company, failures in existing service coverage, or external trends and factors. Large or immature deployments face the need to constantly monitor requests poorly-covered by the existing service for identifying points where the distribution of the input data has changed, for effective and efficient model retraining.

Publicly-available datasets of real-word user requests in customer deployments are extremely scarce due to company agreements, confidentiality and privacy considerations. While the proposed pipeline has been evaluated on large customer deployments, here we create a novel, carefully curated dataset that reliably imitates the characteristics of user requests, and further conduct drift detection evaluation on the collected data.

Intent classifier training examples are inherently designed to reliably represent user requests a VA. Naturalistic user requests, however, typically have several characteristics in which they differ from training examples: (1) the same request semantics can be conveyed in many possible ways, while training examples of the respective intent typically cover the potential diversity only to a partial extent; (2) contrary to intent training examples that only contain unique phrases, actual user requests include many duplicates (multiple customers asking the same question); and (3) customer requests in real-world systems are typically shorter (often significantly so) than classifier training examples.

Using CLINC150 \citep{larson-etal-2019-evaluation}, a large a diverse 150-intent classification dataset, we generate its extended version simulating the nature of customer requests--- CLINC150-SUR (simulated user requests)---by addressing the mentioned distinctions, as detailed below. A typical large customer virtual assistant size varies between few dozens to hundreds of intents, often spanning multiple domains. CLINC150 is multi-domain 150-intent dataset, which makes it a suitable test-bed for our drift detection experiments.

\paragraph{Data Augmentation for Diversity} 
We achieve higher request diversity by applying LAMBADA \citep{anaby2020not}, a tool for classifier training set augmentation; LAMBADA generates phrases sharing the same semantic charge as the seed classification examples provided as input. Next, we apply the \href{https://github.com/PrithivirajDamodaran/Parrot_Paraphraser}{Parrot} paraphrasing framework, generating up to five additional phrasing variants for each data example. While the LAMBADA generates in-class semantic-preserving but lexically-diverse examples, Parrot adheres to more conservative choices by producing slight variations of its input phrases. As an example, considering the CLINC150 "insurance" intent training example "can you tell me the name of my insurance plan?", the request "can you tell me what insurance plan I am signed up for?" was generated by LAMBADA, and "can you tell me what insurance plan I have?" was further added by Parrot.

\begin{figure}[h!]
\centering
\resizebox{1\columnwidth}{!}{
\includegraphics{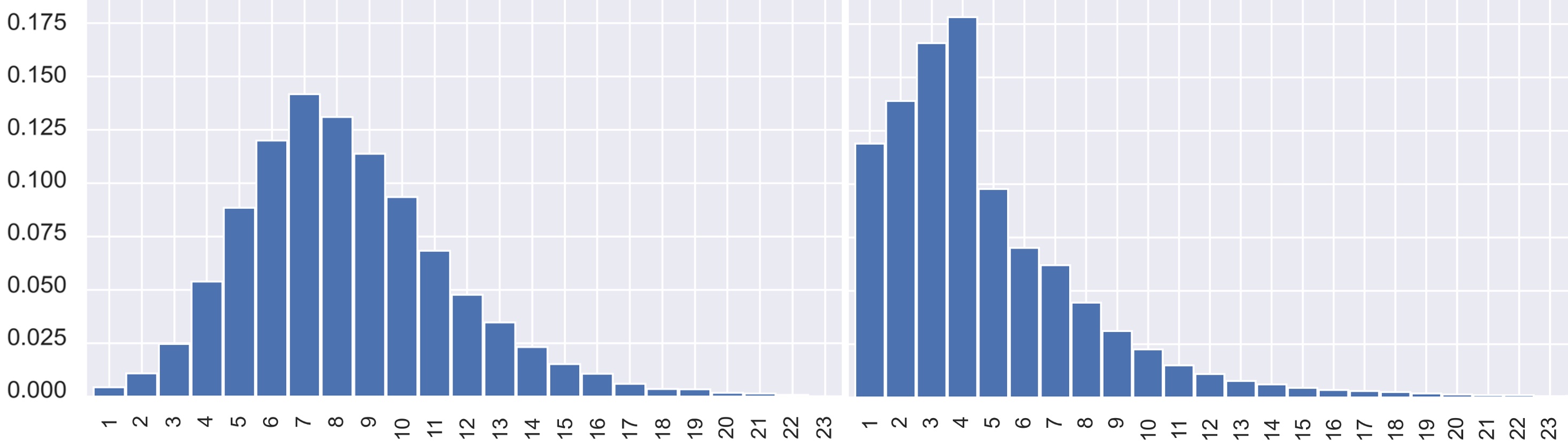}
}
\caption{Distribution of intent classification examples length from \citet{larson-etal-2019-evaluation} (left), and user requests length in a large-scale customer deployment (right). A bar's height mirrors the relative ratio of utterances with a certain number of words in the dataset.}
\label{fig:request-length}
\end{figure}

\paragraph{Weighted Upsampling of Duplicates} 
Figure~\ref{fig:request-length} illustrates the differences in request length: the distribution of the relative ratio of user requests of certain length (in tokens) observed in the intent classification data (left) significantly differs from that evident in a real-world large proprietary dataset of streaming customer requests (right). In particular, over 66\% of actual user utterances consist of up to 5-token requests, while only 18\% of the CLINC150 data exhibit similar length. Short and often non-informative requests challenge tools that rely on textual semantic similarity, hence affecting the process of drift detection. We therefore strive to imitate the naturalistic length-based request distribution in our dataset, by upsampling the augmented data to preserve similar length-based distribution as in Figure~\ref{fig:request-length} (right). As a concrete example, short requests like "insurance" and "my insurance" were upsampled, mirroring their natural frequency in a real-world VA, while only a single instance of "I would like to know all of the covered benefits that are given by my health care plan" remained in the final dataset. We report the CLINC150-SUR statistics in Table~\ref{tbl:dataset-details}. The final collection of $\sim$600K requests is made publicly available.

\begin{table}[hbt]
\centering
\resizebox{\columnwidth}{!}{
\begin{tabular}{l|rr}
dataset & mean & total\\ \hline
original (\citet{larson-etal-2019-evaluation}) & 150.0 & 22.5K \\
\hspace{0.1cm} + generation (\citet{uziel2023}) & 448.4 & 67.3K \\
\hspace{0.1cm} + rephrasing (Parrot)& 1.65K & 250K \\
\hspace{0.1cm} + upsampling & 4K & 600K\\

\end{tabular}
}
\caption{Statistics of the created dataset CLINC150-SUR. Mean number of requests per intent, and the total amount of requests is reported per each expansion step.}
\label{tbl:dataset-details}
\end{table}

\section{Interpretable Drift Detection}
\label{sec:model}

We propose and evaluate a multi-step approach for reliable and interpretable change point detection in textual streams.
The end-to-end pipeline of drift detection and interpretation is illustrated in Figure \ref{fig:process-overview}; 
below we describe each step in more detail.

\subsection{Drift Point Detection}
\label{sec:drift-point-detection}

\paragraph{Model Training} 
The initial distribution of the dataset was learned by training an autoencoder (AE) on a seed dataset representing data distribution at the beginning of monitoring window.\footnote{We used the \href{https://scikit-learn.org/stable/modules/generated/sklearn.neural_network.MLPRegressor.html}{MLPRegressor} implementation at sklearn with three hidden layers of 600, 150 and 600. MLPRegressor functions as an autoencoder when provided with identical input and output representations.} An AE is a special type of neural network that is trained to reproduce its input using the encoder-decoder architecture. Given a dense text representation (embedding) $e$, an AE first encodes the text into a lower-dimensional latent representation, then decodes the latent representation back to the text representation $\hat{e}$; it essentially learns to compress the data while minimizing the reconstruction error $\mathcal{J}(e, \hat{e})$. The network's "success" at reconstructing a new example at inference time reflects the correlation of this instance to the nature of data the model was trained on. In the context of text processing, autoencoders have been effectively applied to the task of anomaly and novelty detection \citep{paula2016deep, zhou2017anomaly, mei2018using}.\footnote{Our future work includes experimenting with variational autoencoder (VAE), introducing a regularisation term into its loss function for better generalization capabilities.} Operating at the individual instance level, an autoencoder detects a pool of "outliers" from within a given data. The Universal Sentence Encoder (USE; \citealt{cer2018universal}) was used for encoding requests into dense representations $e$, due to its runtime efficiency.

Another intuitive approach to instance-level outlier detection employs the \textit{perplexity} metric: the extent of surprisal of a pretrained language model by an unseen text. The approach has been studied by \citet{freeman2021detecting} for detecting anomalies in streams of short texts; we leave the investigation of this alternative approach for future work.

\paragraph{Drift Candidates Detection} Incoming request data is split into fixed-sized batches (in terms of the number of requests), and the model is applied on every new batch as it arrives, computing reconstruction similarity for each request in the chunk. A request's embedding $e$ reconstruction similarity is computed as the $cosine$ similarity of its original representation to the representation of its reconstructed counterpart $\hat{e}$, i.e., $cosine(e, \hat{e})$. Utterances poorly reconstructed by the anchor model are considered \textit{outliers}: requests where $cosine(e, \hat{e}){<}\gamma$, for a predefined $\gamma$, are stored in the outlier pool $\mathcal{O}$. For data chunk at the time step $t_i$, its similarity $s_i$ to the anchor dataset is calculated, producing a growing sequence of numerical observations $\mathcal{S}{=}\{s_1, s_2, \dots, s_{t-1}, s_t\}$; at each time step $t$, the sequence is passed to change-point detection module.

\paragraph{Change-Point Detection}
Drift is indicated by a change-point in the distributions of the observed similarities $\mathcal{S}$. That is, index $t_s<t$ is a change-point, the true starting index of the change, if the distributions of values $\{s_1,\dots,s_{t_s-1}\}$ (before) and $\{s_{t_s},\dots,s_t\}$ (after) differ significantly. We apply the change-point model (CPM; \citealt{doi:10.1080/00224065.2012.11917887}, implemented in \textrm{R} as \textrm{cpm}, \citealt{ross2015cpmpackage}) algorithm repeatedly on the past $\{s_1,\dots,s_t\}$ at each $t$. That is, at a given $t$, if the most significant candidate split point is significant enough (the CPM p-value is a fixed $<\alpha$, say 0.05), we say that $t_d=t$ is the detection index; moreover, the most significant split index, $t_p$\:---where $t_p<t_d$\:--- is $\hat{t}_s$, that is, the best guess of the actual change point $t_s$, if it actually happened.  The CPM is unique in that it correctly maintains the false positive rate at $\alpha$ (e.g., 0.05) even though it is applied repeatedly in sequence, testing each potential change location for each $t$.  That is, if a detection is made at $t$, the probability the detection was false (there was no change point)  is at most $\alpha$.  Furthermore, this method makes no parametric assumptions about the distributions.  See \citet{ackerman2020detection} for further details and discussion.

\subsection{Drift Interpretation}
The predicted change point $t_p$ detected over a sequence of similarity observations $\mathcal{S}$ is further used as an indicator for the start point of topical novelties; all outlier utterances from the outlier pool $\mathcal{O}$ that occurred after the predicted change point (with time indices $t\in[t_p,\:t_d]$) are utilized for semantic grouping, or clustering. In our scenario, an effective clustering procedure should have several properties. First, the number of clusters is unknown, and has to be discovered by the clustering algorithm. Second, the nature of data typically implies several large and coherent clusters, where users repeatedly introduce very similar requests, and a very long tail of unique (often noisy) requests that do not have similar counterparts. We apply the RBC clustering approach used by \citet{rabinovich2022gaining}, that was specifically tailored for the scenario of unhandled requests in task-oriented dialog systems: the procedure does not require a predefined number of clusters, tolerating non-clusterable instances.

\begin{figure}
\centering
\resizebox{\columnwidth}{!}{
\includegraphics{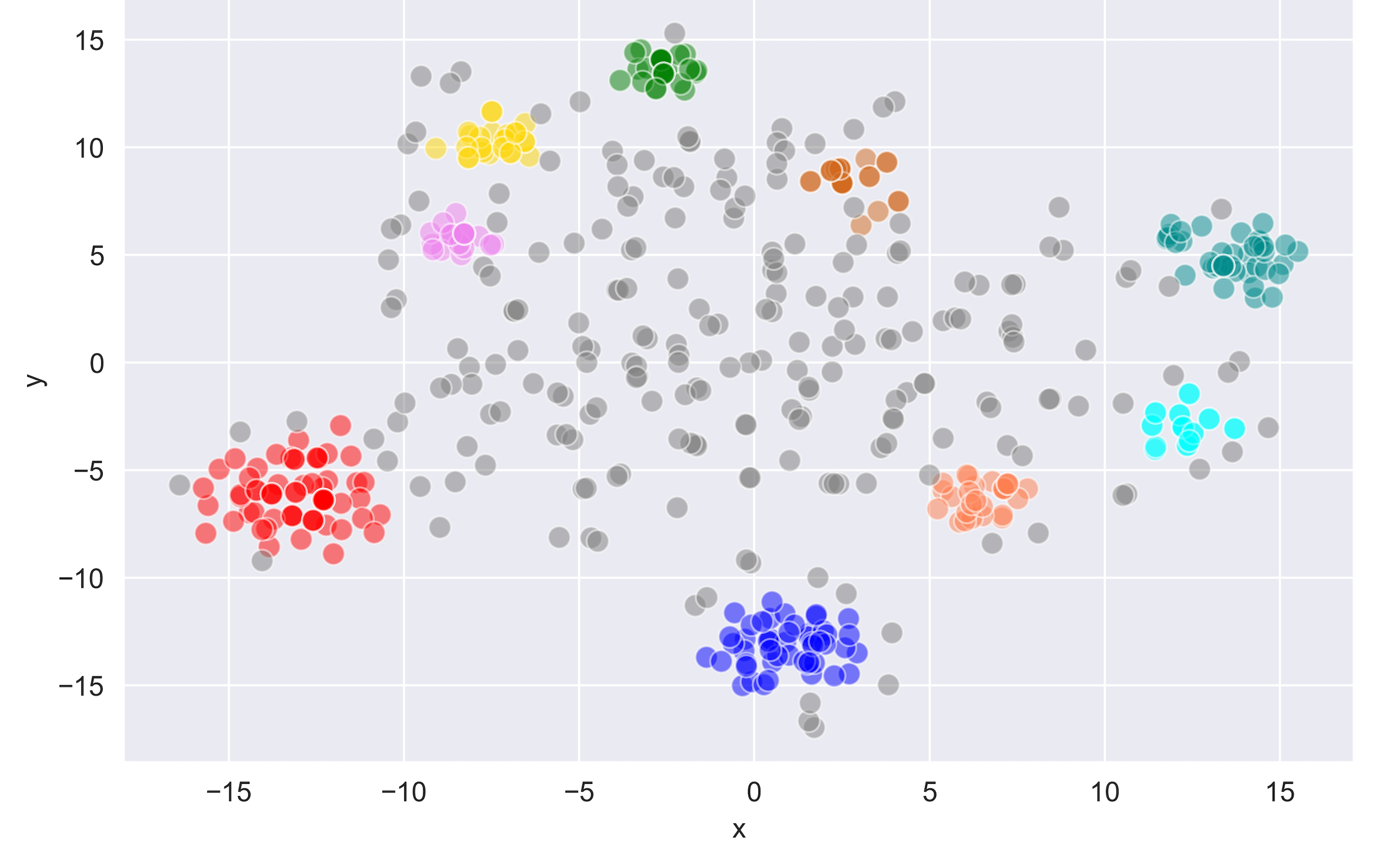}
}
\caption{t-SNE projection of a sample of outlier user requests in a production task-oriented dialog system. Identified clusters are in color, instances that do not firm up large enough clusters -- in grey.}
\label{fig:clustering-space}
\end{figure}

Figure~\ref{fig:clustering-space} illustrates a typical outcome of the clustering process; identified clusters---each representing likely instances of the same potentially novel intent---are depicted in color, while non-clusterable instances, constituting approximately half of the instances, appear in grey.

\section{Experimental Results}

In this section we describe the experimental setup and results for two major evaluation phases: change-point detection and drift interpretation.

\subsection{Drift Point Detection Results}

\subsubsection{Experimental Setup}

\begin{figure}[h!]
\centering
\resizebox{1.0\columnwidth}{!}{
\includegraphics{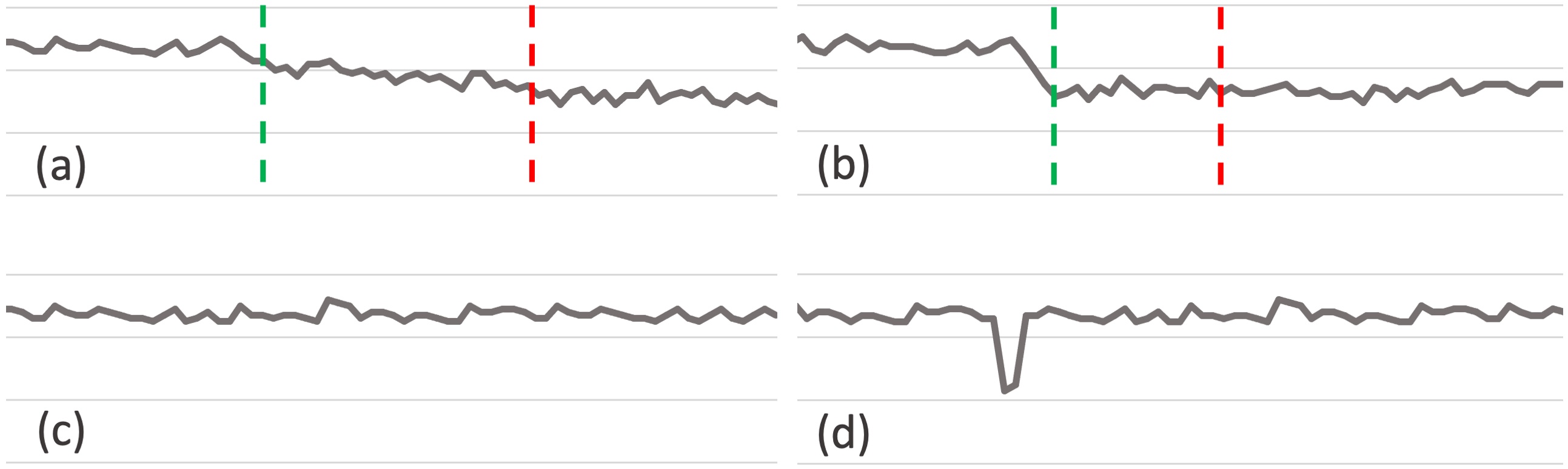}
}
\caption{Illustration of data batch similarity signal to the anchor model in various drift scenarios: (a) gradual, (b) uniform, (c) no drift, (d) short-lived anomaly. Note the slight signal decrease in (a) and (b); the dashed green line denotes the actual drift point, the dashed red line denotes the potential detection point.}
\label{fig:drift-types}
\end{figure}

\paragraph{Drift Scenarios} 
Introducing drift into a VA request stream for thorough evaluation is a non-trivial task. A realistic setup would entail simulating one or more novel (unseen) topics that gradually comprise an increasing number of requests over time, as in the case of a new feature introduced by a service provider, being gradually adopted by customers. Another plausible scenario is where novel topics are introduced by service interrupt or unexpected failure; in that case, one may expect a steep increase in atypical requests, followed by nearly plateau distribution over time. We refer to these scenarios as (a) and (b), respectively. Correctly identifying scenarios where no drift was introduced is of considerable importance as well, ensuring the system is not prone to false positives. We cover this scenario by two additional experimental setups where (c) no drift is introduced, and (d) a short-lived anomaly is introduced spanning a small number of consecutive data batches. The various drift scenarios, as reflected in data batch similarities $\mathcal{S}{=}\{s_1, s_2, \dots, s_{t-1}, s_t\}$ to the anchor model, are depicted in Figure~\ref{fig:drift-types}.

\paragraph{Drift Detection}
Multiple setup decisions were used for drift detection experiments in this study. A series of 32 temporally-ordered data batches of 5K requests each was generated from shuffled request dataset CLINC150-SUR, where $\sim${5}\% randomly selected intents out of 150 were held out for drift injection. Although drift detection is likely to be more robust on large data chunks, we found consistent behavior when working with batches containing as few as a couple of hundreds of requests.\footnote{The precise definition of data chunk size varies between deployments, and depends on a system traffic, the number of intents and drift detection tolerance threshold, among others.} The first two data chunks where used for training the anchor AE model, and the remaining 30 for simulating a temporal stream of requests.\footnote{30 observations where found sufficient for effective detection of drift in our experiments; consistent behavior was observed when increasing the temporal stream length.} Drift was introduced by adding requests from held-out intents either gradually (a) or uniformly (b) starting at the middle of the stream, i.e., at time step $t{=}$15. The novel intents causing the drift, as well as the subset of requests spanning the 32 data chunks, were selected at random per every experiment.

A single parameter that was tuned for drift detection is the outlier detection threshold $\gamma$ (see Section~\ref{sec:drift-point-detection}). While typically lying within the [0.75, 1] range, the optimal value of $\gamma$ varies according to the data nature. In multi-domain deployments, higher $\gamma$ would typically tolerate the inherent diversity of the incoming requests; in contrast, in single-domain deployments, lower $\gamma$ would impose a stricter threshold for outliers detection. We tune $\gamma$ to optimize detection accuracy per deployment using a held-out portion of data; for the CLINC150-SUR used in this work, $\gamma$ was set to 0.775.

\subsubsection{Experimental Results}

\begin{table*}[h!]
\centering
\resizebox{\textwidth}{!}{
\begin{tabular}{l|c|cc|ccc|ccc}
metric & \splitcell{instance\\ level} & \multicolumn{2}{c|}{\splitcell{false\\ detection rate}} & \multicolumn{3}{c|}{drift type: gradual} & \multicolumn{3}{c}{drift type: uniform}\\ \hline
& & \hspace{0.2cm}FP & FN & \splitcell{detection\\ offset} & \splitcell{detection\\ deviation} & \splitcell{drift rate at\\ detection} & \splitcell{detection\\ offset} & \splitcell{detection\\ deviation} & \splitcell{drift rate at\\ detection}  \\ \hline
AE (our approach) & \cmark & 0.04 & 0.04 & 7.08 & 2.03 & 0.017 & 5.82 & 1.07 & 0.022 \\ \hline
IRPR \cite{zhao-etal-2017-learning} & \xmark & 0.08 & 0.04 & 7.02 & 1.79 & 0.015 & 5.71 & 1.01 & 0.020 \\
Medoid \cite{article} & \xmark & 0.10 & 0.21 & 9.17 & 5.08 & 0.017 & 7.23 & 1.25 & 0.020 \\
MAUVE \cite{NEURIPS2021_260c2432} & \xmark & 0.15 & 0.11 & 7.02 & 3.05 & 0.023 & 5.91 & 1.09 & 0.021 \\
FID \cite{NIPS2017_8a1d6947} & \xmark & 0.09 & 0.13 & 8.22 & 4.16 & 0.026 & 6.28 & 1.01 & 0.024 \\

\end{tabular}
}
\caption{Drift detection evaluation results; the lower, the better. 
Mean results over 100 experiments are reported, where false detection rate is averaged over the gradual and uniform scenarios. AE is the only approach that operates at the instance-level out-of-the-box. FN are averaged over gradual and uniform scenarios.}
\label{tbl:similarity-metrics-evaluation}
\end{table*}

We compare the performance of the AE-based similarity measurements to various dataset similarity metrics suggested in literature. While computing anchor-batch similarity, metrics operating at the \textit{instance} level (i.e., aggregating the similarity of individual requests to the anchor model) can be used for seamless generation of the outlier pool, further used for drift interpretation. Metrics that operate at the \textit{batch} level typically make use of measures of central tendency and dispersion, comprising a less natural (albeit adaptable) choice for our scenario.

\paragraph{Dataset Similarity Metrics}
A comprehensive evaluation of various dataset semantic similarity metrics has been conducted recently by \citet{kour-etal-2022-measuring}. We evaluate our approach against several metrics from that work. IRPR \cite{zhao-etal-2017-learning} is a corpus distance metric based on information-retrieval techniques focused on precision and recall. Medoid \cite{article} applies cosine similarity over the arithmetic mean of embeddings of two textual sets. MAUVE \cite{NEURIPS2021_260c2432} estimates the gap between two texts using KL-divergence over the area under the information divergence frontiers. FID \cite{NIPS2017_8a1d6947} calculates the 2-Wasserstein distance on fitted continuous multivariate Gaussian over two datasets.

\paragraph{Drift Detection Metrics} In line with previous work on drift detection \citep{wang2018real, ackerman2021automatically} we report multiple results: \textit{detection offset} denotes the number of steps between the two points ($t_s$, $t_d$): where the drift was introduced ($t_s$) and detected ($t_d$), i.e., |$t_d{-}t_s$|; \textit{detection deviation} measures the difference between the actual drift injection point ($t_s$), and the drift point suggested by the CPM module ($t_p{=}\hat{t}_s$), i.e., |$\hat{t}_s{-}t_s$|; finally, \textit{drift rate at detection} denotes the relative rate of drift requests within the entire amount of requests in the batch corresponding to the detection point $t_d$. We report these metrics for both gradual (a) and uniform (b) drift scenarios in Figure~\ref{fig:drift-types}. An additional measurement of interest is the \textit{false negative} (FN) rate, the proportion of experiments where drift was not detected over the series of 30 observations, despite drift that was injected.

We address the two no-drift scenarios---(c) and (d)---in Figure ~\ref{fig:drift-types}, by reporting the rate of \textit{false positives} (FP): the proportion of experiments where drift was erroneously detected by the CPM module using each one of the similarity metrics.

Table~\ref{tbl:similarity-metrics-evaluation} reports the results. Using AE for computing datasets similarity performs roughly on par with IRPR, and outperforms other approaches, across the board. Detection offset and deviation are higher in the gradual drift scenario (7.08 and 2.03 vs. 5.82 and 1.07, respectively) reflecting the more challenging setup of a growing drift, compared to the stable plateau drift spread starting a certain point. The relatively low drift rate at detection (1.7\%--2.2\%) implies that the procedure is sensitive to drift at its early stage. On the other hand, the low rate of false positive and negatives is indicative of the robustness of the detection routine. 

Finally, in the next paragraph, we show that our approach outperforms the model-dependent approach leveraging a classifier confidence scores.

\paragraph{Model-dependent Experiments}
Additional set of experiments was conducted using intent classifier posterior estimates as indicator for data drift. Identical experimental setup was applied, where 95\% of intent training set from CLINC150 (seed data) was used for training SVM classifier with training instances' embeddings. Roughly 5\% of intents were randomly selected and held-out as "novel" at each experiment. The pretrained classifier was then used to classify batches of requests from CLINC150-SUR corresponding to the seed data (before injecting drift), and to the seed data extended with drift requests (starting the drift point). Mean classifier confidence was computed for every request batch yielding a series of observations, which was further inut to the change-point detection module (see Section~\ref{sec:drift-point-detection}).

Drift was detected in less than 20\% of the experiments, compared the 96\% with the model-agnostic approach, highlighting the benefits of the direct access to data for drift detection.

\subsection{Drift Interpretation Results}

Drift interpretation is a two-step process: first, outlier requests are grouped together based on their semantics, thereby, firming up dense clusters conveying the same intent; second, identified clusters are assigned with names for better consumability.

The subset of outlier requests $\mathcal{O}$ starting from the predicted drift point $\hat{t}_s$ is further used for identifying novel topics, indicated by dense clusters of requests that share similar semantics. The pool of outliers is not limited to novelties but also contains requests pertaining to existing ("known") intents that could not be successfully reconstructed by the anchor model, and requests that pertain to topics that were left out of the VA scope by design. 

We apply the RBC clustering approach by \citet{rabinovich2022gaining} with defaults for surfacing topical clusters, and focus on topical coverage (recall) in our evaluation. Each cluster---a group of similar outlier utterances---is assigned an intent label based on the majority of its members, and the coverage is computed as the ratio of detected intents out of injected drift intents for each individual experiment. The mean recall in 100 experiments was 0.709, meaning that on average, 70\% of the injected drift intents were identified as such.

Inspecting names (automatically) assigned by the algorithm to detected outlier clusters, we can identify a significant degree of overlap between those names and drift intent labels, which were presumably created by human annotators. As an example, the "exchange\_rate" drift intent was identified as such and assigned the label "exchange rate" by the clustering algorithm; requests from the "order\_status" drift intent were named as "order tracking", and "redeem\_rewards" was surfaced as "redeem rewards points".

\section{Conclusions}
We propose and evaluate a pipeline for model-agnostic change-point detection in the context of drift detection in task-oriented dialog systems NLU module. We demonstrate the benefits of the proposed approach on an expanded version of an intent classification training dataset, that closely imitates the nature of streaming requests towards a task-oriented dialog system -- the dataset that we make available to the research community.
We demonstrate that AE can be used for effective and efficient change-point detection, performing on par with state-of-the-art dataset similarity metrics, while operating at the instance level.

Our future directions include experimenting with the (baseline) metric of language model perplexity as well as variational autoencoders for the task of drift detection in streams of short texts. Extending experimental setup to the multi-lingual setting is another direction we plan to pursue.

\section*{Acknowledgements}
We are grateful to the three anonymous reviewers for their constructive feedback. We would also like to thank Orna Raz, Eitan Farchi and Haode Qi for their kind help and fruitful discussions.

\bibliography{custom}
\bibliographystyle{acl_natbib}


\end{document}